\documentclass{article}

\usepackage[preprint]{neurips_2025}

\usepackage[utf8]{inputenc} 
\usepackage[T1]{fontenc}    
\usepackage{hyperref}       
\usepackage{url}            
\usepackage{booktabs}       
\usepackage{amsfonts}       
\usepackage{nicefrac}       
\usepackage{microtype}      
\usepackage{xcolor}         
\usepackage{amsmath}
\usepackage{amssymb}
\usepackage{amsthm}
\usepackage{graphicx}
\usepackage{float}

\usepackage{enumitem}
\usepackage{subcaption}
\usepackage{tikz}
\usetikzlibrary{automata, positioning, arrows.meta, calc}

\title{Neural Networks as Universal Finite-State Machines: A Constructive Deterministic Finite Automaton Theory}

\author{%
  Sahil Rajesh Dhayalkar
  \\
  Brain Corporation \\
  San Diego, CA\\
  \texttt{sahil.dhayalkar@braincorp.com} \\
}

\begin{document}

\maketitle

\begin{abstract}
We present a complete theoretical and empirical framework establishing feedforward neural networks as universal finite-state machines (N-FSMs). Our results prove that finite-depth ReLU and threshold networks can exactly simulate deterministic finite automata (DFAs) by unrolling state transitions into depth-wise neural layers, with formal characterizations of required depth, width, and state compression. We demonstrate that DFA transitions are linearly separable, binary threshold activations allow exponential compression, and Myhill-Nerode equivalence classes can be embedded into continuous latent spaces while preserving separability. We also formalize the expressivity boundary: fixed-depth feedforward networks cannot recognize non-regular languages requiring unbounded memory. Unlike prior heuristic or probing-based studies, we provide constructive proofs and design explicit DFA-unrolled neural architectures that empirically validate every claim. Our results bridge deep learning, automata theory, and neural-symbolic computation, offering a rigorous blueprint for how discrete symbolic processes can be realized in continuous neural systems.
\end{abstract}

\section{Introduction}

Neural networks are celebrated for their universal approximation capabilities \cite{hornikmultilayer, cybenko1989approximation}, yet the precise nature of their computational expressivity remains an active area of theoretical inquiry. While it is well-understood that neural networks can approximate any continuous function under suitable conditions, less is known about how such networks relate to classical models of computation, such as finite-state machines (FSMs) and automata. This paper presents the first comprehensive theoretical and empirical framework establishing \emph{neural networks as universal finite-state machines} (N-FSMs), with a constructive and exact simulation of \emph{deterministic finite automata} (DFAs), thereby formally bridging deep learning and automata theory.

Our core contribution is a set of rigorous theorems, lemmas, and corollaries demonstrating that feedforward neural networks with ReLU or threshold activations can exactly emulate the behavior of deterministic finite automata (DFAs) over finite-length strings. Unlike traditional universal approximation theorems, our results capture bounded-memory computation and deterministic transition dynamics, offering a new lens through which to analyze the symbolic computation capabilities of neural architectures. We also provide a constructive theory of DFA emulation by explicitly ``unrolling'' automaton transitions into depth-wise neural layers, a methodology previously unexplored in the neural-symbolic literature.

To complement the theoretical results, we design a suite of DFA-unrolled neural architectures that empirically validate each claim across increasingly complex properties: from direct DFA simulation, linear separability of transitions, and compact binary state encodings, to equivalence class embedding and information-theoretic compression. We further delineate the expressive boundary of N-FSMs, demonstrating empirically that fixed-depth feedforward networks cannot recognize context-free languages requiring unbounded memory, such as $\{a^n b^n\}$, thereby aligning our results with the Chomsky hierarchy \cite{chomskyhierarchy, introautomata}.

Our findings contribute to both neural theory and symbolic AI by:
\begin{itemize}[itemsep=0pt, topsep=0pt]
    \item Establishing formal proofs that DFAs can be exactly simulated by finite-depth neural networks, and identifying the architectural depth and width requirements.
    \item Demonstrating that neural networks can learn Myhill-Nerode equivalence class representations, providing a geometric interpretation of regular languages in latent space.
    \item Showing that DFA state encodings can be compressed to $\mathcal{O}(\log n)$-dimensional embeddings while preserving class distinction.
    \item Empirically validating the expressivity limits of N-FSMs through DFA-unrolled experiments.
\end{itemize}

This work advances the field of neural-symbolic computation \cite{besold2017neuralsymboliclearningreasoningsurvey, garcez2019neuralsymboliccomputingeffectivemethodology} by formalizing how neural networks can simulate deterministic symbolic processes and by providing a blueprint for designing neural architectures that integrate classical computation principles. In doing so, it lays the foundation for future research on efficient neural representations of symbolic systems, differentiable automata, and the theoretical limits of neural computation. The paper uses a structure with theorems, lemmas, and corollaries, similar to prior work~\cite{zhang2015learninghalfspacesneuralnetworks, hanin2018approximatingcontinuousfunctionsrelu, neyshabur2015searchrealinductivebias, dhayalkar2025combinatorialtheorydropoutsubnetworks, dhayalkar2025geometryrelunetworksrelu}.

\section{Related Work}

\textbf{Neural-Symbolic Integration and Computability:} Integrating neural computation with symbolic reasoning has long been a goal of neural-symbolic computing \cite{besold2017neuralsymboliclearningreasoningsurvey, garcez2019neuralsymboliccomputingeffectivemethodology}. Early efforts combined symbolic logic with connectionist models using recurrent or hybrid architectures, but lacked formal characterizations of computational equivalence. While universal approximation theorems \cite{cybenko1989approximation, hornikmultilayer} established that feedforward networks can approximate any continuous function, they address function approximation in the limit and not exact simulation of discrete models like finite-state machines (FSMs). Later work examined the Turing completeness of recurrent and deep networks \cite{siegelmann1995computational}, yet finite-state computation in fixed-depth feedforward networks remains underexplored.

\textbf{Finite-State Machines and Boolean Circuits in Neural Networks:} Empirical studies have shown that recurrent networks and transformers can implicitly learn finite automata representations when trained on regular languages \cite{weiss2018practical, rabinovich2017abstract}. These works used black-box probing techniques but did not offer formal proofs or constructive architectures guaranteeing FSM simulation. Separately, threshold units simulating Boolean logic have been foundational in circuit complexity theory \cite{hastad1986almost} and prior neural-symbolic work applied Boolean gates for propositional reasoning \cite{garcez2019neuralsymboliccomputingeffectivemethodology}. However, no previous work linked these to DFA transition simulation or the compression advantages of binary state encodings. Our results unify these areas, providing formal proofs that binary threshold networks can simulate DFA transitions exactly with logarithmic compression.

\textbf{Expressivity Limits and Language Hierarchies:} Discussions of neural network expressivity often invoke the Chomsky hierarchy \cite{chomskyhierarchy}. While fixed-depth feedforward networks can model regular languages, they lack the memory capacity to recognize non-regular languages like context-free languages requiring unbounded memory \cite{introautomata}. Our Theorem \hyperref[theorem_3]{3} and Corollary \hyperref[corollary_3_1]{3.1} rigorously formalize this limitation and validate the boundary between regular and non-regular language recognition.

\textbf{Geometric Representations and Neural Automata:} Some studies have explored geometric representations of language structure and compositionality in neural networks. Hupkes et al. \cite{hupkes2020compositionality} analyzed compositional generalization, and Weiss et al. \cite{weiss2018practical} observed automaton-equivalent state embeddings in recurrent networks. However, these lacked formal guarantees. In contrast, our Theorem \hyperref[theorem_2]{2} and Corollary \hyperref[corollary_2_1]{2.1} provide the first formal and empirical demonstration that Myhill-Nerode equivalence classes can be represented as distinct geometric embeddings with provable compression. Concurrent work has also explored combining automaton structures with neural architectures to enhance interpretability and generalization \cite{graves2016adaptive, hahn2020theoretical}, but these approaches remain empirical or limited to theoretical bounds without constructive architectural proofs.

\textbf{Summary:} Prior efforts have relied on empirical observations, heuristics, or probing-based approaches. In contrast, our work presents a complete theoretical and empirical framework proving that feedforward neural networks can serve as universal finite-state machines, bridging foundational gaps between deep learning, automata theory, and neural-symbolic computation.

\section{Theoretical Framework}

\subsection{Preliminaries: Deterministic Finite Automata}
A Deterministic Finite Automaton (DFA)is a tuple $\mathcal{A} = (Q, \Sigma, \delta, q_0, F)$ where:
\begin{itemize}
    \item $Q$ is a finite set of states, $|Q| = n$.
    \item $\Sigma$ is a finite input alphabet, $|\Sigma| = k$.
    \item $\delta : Q \times \Sigma \to Q$ is the transition function.
    \item $q_0 \in Q$ is the initial state.
    \item $F \subseteq Q$ is the set of accepting states.
\end{itemize}

Given a finite-length input string $x = (s_1, s_2, \dots, s_T) \in \Sigma^T$, the DFA reads one symbol at a time and updates its state according to the transition function $\delta$. The extended transition function $\hat{\delta}$ maps strings to final states by recursively applying $\delta$:
\[
\hat{\delta}(q_0, x) = \delta(\dots \delta(\delta(q_0, s_1), s_2), \dots, s_T).
\]
We write $\hat{\delta}(x)$ to denote the final state reached when starting at $q_0$ and reading string $x$.

The DFA \emph{accepts} $x$ if $\hat{\delta}(x) \in F$. The corresponding \emph{language indicator function} is defined as:
\[
\phi(x) = 
\begin{cases}
1, & \text{if } x \text{ is accepted by } \mathcal{A}, \\
0, & \text{otherwise}.
\end{cases}
\]

The \emph{language recognized} by $\mathcal{A}$ is the set:
\[
L(\mathcal{A}) = \{ x \in \Sigma^* \mid \phi(x) = 1 \} = \{ x \in \Sigma^* \mid \hat{\delta}(x) \in F \}.
\]

This formalism underpins all theorems, lemmas, and corollaries that follow.

\subsection{Theorem 1: Existence of FSM-Emulating Neural Networks}
\label{theorem_1}
Let $\mathcal{A} = (Q, \Sigma, \delta, q_0, F)$ be a deterministic finite automaton (DFA), where $|Q| = n$ and $|\Sigma| = k$. For any finite input length $T \in \mathbb{N}$, there exists a feedforward neural network $f_\theta: \mathbb{R}^{Tk} \rightarrow \mathbb{R}$ such that for every input string $x = (s_1, s_2, \dots, s_T) \in \Sigma^T$, the network outputs
\[
f_\theta(x) = 
\begin{cases}
1 & \text{if } \mathcal{A} \text{ accepts } x, \\
0 & \text{otherwise},
\end{cases}
\]
i.e., $f_\theta(x) = \phi(x)$, where $\phi: \Sigma^T \rightarrow \{0,1\}$ is the language indicator function of $\mathcal{A}$.
(Proof provided in Appendix \hyperref[proof_theorem_1]{A.1})

\begin{figure}[h]
    \centering
    \includegraphics[width=\textwidth]{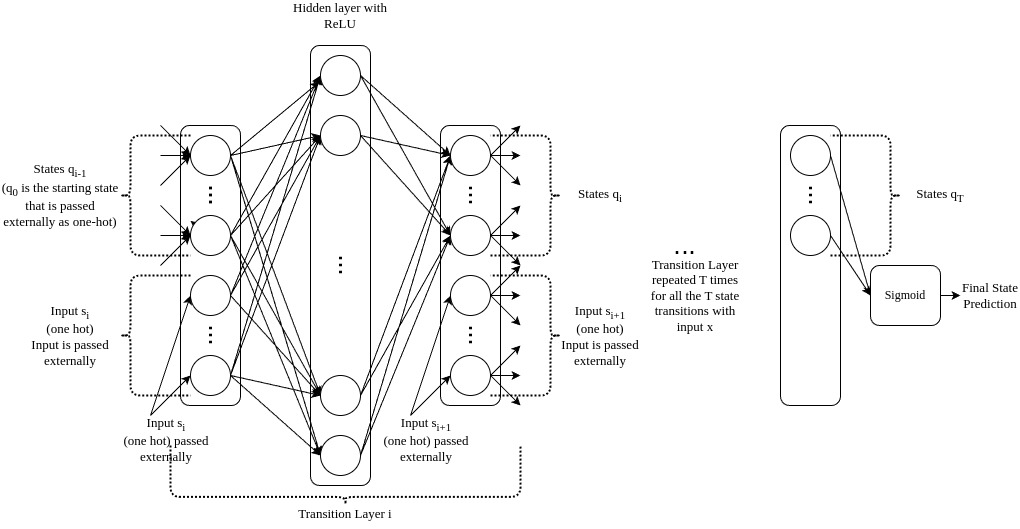}
    \caption{Figure depicting a Neural Network emulating a DFA}
    \label{figure_nn_as_fsm}
    \label{fig:theorem_1}
\end{figure}

This theorem asserts that any DFA over finite-length strings can be exactly emulated by a feedforward neural network (FNN) without recurrence or explicit memory. This goes beyond the universal approximation theorem by focusing on bounded memory and deterministic computation, and opens a new direction in understanding how classical computation models (finite automata) can be encoded into the structure of neural networks. Figure~\ref{figure_nn_as_fsm} shows how a neural network would emulate a DFA.

The core idea is that each layer of the network simulates one transition step of the automaton, propagating the current state forward based on the current input symbol. The network "unrolls" the automaton in depth, with each layer encoding a single application of the transition function $\delta$.


\subsection{Lemma 1: Linear Separability of State Transitions}
Let $\mathcal{A} = (Q, \Sigma, \delta, q_0, F)$ be a DFA, where each state $q_i \in Q$ and symbol $s_j \in \Sigma$ are encoded as one-hot vectors $e_i \in \mathbb{R}^{n}$ and $u_j \in \mathbb{R}^{k}$ respectively. Then there exists a two-layer feedforward neural network with ReLU activations that computes the transition function
\[
g: \mathbb{R}^{n+k} \rightarrow \mathbb{R}^{n}, \quad g([e_i; u_j]) = e_{\delta(q_i, s_j)}.
\]
(Proof provided in Appendix \hyperref[proof_lemma_1]{A.2})

This lemma formalizes the intuition that, because a DFA has a finite transition table, its behavior can be captured by a small neural module. Using one-hot encoding, every valid input pair \(([q_i], [s_j])\) maps to a unique one-hot output vector \([q']\). Since both the domain and codomain are finite and discrete, the mapping is trivially realizable by a ReLU MLP.

This lemma implies that the DFA’s logic is linearly separable in the input space of concatenated state and symbol vectors. Every transition rule \(\delta(q_i, s_j) = q_k\) corresponds to a point in input space that maps to a unique output. The ReLU MLP simply partitions input space into these regions.

Although DFA transition tables are finite and discrete, the claim and construction that ReLU MLPs can realize them using linear separability over concatenated one-hot vectors is new. Prior work does not explicitly connect DFA transitions to linear separability in neural space.





\subsection{Lemma 2: State Encodings via Binary Neurons}
\label{lemma_2}
Let $\mathcal{A} = (Q, \Sigma, \delta, q_0, F)$ be a DFA with $|Q| = n$. Suppose each state $q_i \in Q$ is encoded using a unique binary vector $b_i \in \{0,1\}^{\lceil \log_2 n \rceil}$. Then there exists a feedforward neural network with binary threshold activations (e.g., hard step or sign units) that computes the transition function
\[
g: \{0,1\}^{\lceil \log_2 n \rceil + k} \rightarrow \{0,1\}^{\lceil \log_2 n \rceil}, \quad g([b_i; u_j]) = b_{\delta(q_i, s_j)},
\]
where $u_j$ is a one-hot encoding of $s_j \in \Sigma$. (Proof provided in Appendix \hyperref[proof_lemma_2]{A.3})

This lemma establishes that DFA transitions can be simulated using exponentially compact binary encodings and binary threshold activations, replacing one-hot encodings which are large and sparse. By representing states with $\lceil \log_2 n \rceil$-bit vectors and using step or sign activations, the DFA transition function becomes equivalent to a Boolean circuit—a key theoretical connection not previously formalized in neural-symbolic literature. Since binary encoding is exponentially more compact, it is useful for implementations with restricted memory or to analyze theoretical compression properties of FSM simulators.

\subsection{Theorem 2: FSM Equivalence Class Representation}
\label{theorem_2}
Let $\mathcal{A} = (Q, \Sigma, \delta, q_0, F)$ be a DFA with language $\mathcal{L} = L(\mathcal{A})$, and let $\equiv_\mathcal{L}$ be the Myhill-Nerode equivalence relation on $\Sigma^*$. Then there exists a feedforward neural network $f_\theta$ 
such that for all strings $x, y \in \Sigma^{\leq T}$:
\[
x \equiv_\mathcal{L} y \ \Rightarrow \ f_\theta(x) = f_\theta(y),
\]
and
\[
x \not\equiv_\mathcal{L} y \ \Rightarrow \ f_\theta(x) \ne f_\theta(y),
\]
i.e., $f_\theta$ learns an injective mapping from equivalence classes of $\mathcal{L}$ to a vector space. (Proof provided in Appendix \hyperref[proof_theorem_2]{A.4})

The Myhill-Nerode theorem tells us that every regular language partitions the set of strings into finitely many equivalence classes. Each class represents a set of strings that are indistinguishable with respect to future continuations. This theorem shows that a neural network can learn to represent each class with a distinct vector, thus behaving as a classifier over equivalence classes, rather than over raw strings.

This result gives a geometric representation of regular language structure. Each equivalence class corresponds to a unique “state vector” in the output space. Two strings are mapped to the same vector if and only if they lead the DFA to the same state—hence are in the same class.

\subsection{Corollary 2.1: FSM State Compression via Embedding}
\label{corollary_2_1}
Let $\mathcal{A} = (Q, \Sigma, \delta, q_0, F)$ be a DFA with $|Q| = n$ states and minimal with respect to language $\mathcal{L}$. Then there exists a feedforward neural network $f_\theta: \Sigma^{\leq T} \rightarrow \mathbb{R}^d$ with $d = O(\log n)$ such that for all $x, y \in \Sigma^{\leq T}$:
\[
\hat{\delta}(x) = \hat{\delta}(y) \ \Rightarrow \ f_\theta(x) = f_\theta(y),
\]
and
\[
\hat{\delta}(x) \ne \hat{\delta}(y) \ \Rightarrow \ \|f_\theta(x) - f_\theta(y)\| > \epsilon,
\]
for some $\epsilon > 0$. That is, $f_\theta$ maps DFA states into a compressed, continuous latent space that preserves their distinction. (Proof provided in Appendix \hyperref[proof_corollary_2_1]{A.5})

This corollary shows that DFA states, traditionally represented as high-dimensional one-hot vectors, can be embedded into a low-dimensional latent space while preserving their distinguishability. Neural networks thus achieve logarithmic compression of the state space, enabling compact state signatures that reflect automaton-level equivalence, and connects neural representations with the structure of regular languages.

This result leverages the Johnson-Lindenstrauss lemma \cite{Johnson1984ExtensionsOL} to compress DFA state space into an $O(\log n)$-dimensional space while preserving class distinction. This connection between automata theory and latent space compression in neural networks is original





\subsection{Theorem 3: Hardness of Non-FSM Recognition}
\label{theorem_3}
Let $f_\theta$ be any feedforward neural network of finite depth and width, with input $x \in \Sigma^T$ encoded via one-hot vectors. Then there exists a context-free language $\mathcal{L}_{CF} \subset \Sigma^*$ (e.g., $\{a^n b^n \mid n \geq 1\}$) such that no such $f_\theta$ can compute the exact indicator function $\phi_{CF}(x) = \mathbb{I}[x \in \mathcal{L}_{CF}]$ for unbounded input lengths $T$. (Proof provided in Appendix \hyperref[proof_theorem_3]{A.6})

This theorem formally states the limits of expressivity of feedforward neural networks with fixed architecture. While DFAs (regular languages) can be simulated exactly and compactly, higher-level languages in the Chomsky hierarchy \cite{chomskyhierarchy}, such as context-free languages, require unbounded memory, which FNNs fundamentally lack. For example, the language \(\{a^n b^n\}\) cannot be recognized for arbitrarily large \(n\) without mechanisms like a stack or recurrence.

\subsection{Corollary 3.1: Boundary of FSM-Expressive Neural Networks}
\label{corollary_3_1}
Let $f_\theta$ be a feedforward neural network with fixed depth and width. Then:
\begin{enumerate}
    \item There exists a class of regular languages $\mathcal{L}_{\text{reg}}$ such that for every $\mathcal{L} \in \mathcal{L}_{\text{reg}}$, there exists a network $f_\theta$ that exactly computes the membership function $\phi(x) = \mathbb{I}[x \in \mathcal{L}]$.
    \item There exists a disjoint class of non-regular languages $\mathcal{L}_{\text{non-reg}}$ (e.g., context-free or context-sensitive languages) such that for every $\mathcal{L} \in \mathcal{L}_{\text{non-reg}}$, no such network $f_\theta$ can compute $\phi(x)$ for all $x \in \Sigma^*$.
\end{enumerate}

(Proof provided in Appendix \hyperref[proof_corollary_3_1]{A.7})

This corollary formally demarcates the boundary between languages that can and cannot be simulated by neural finite-state machines (N-FSMs). It highlights that N-FSMs are strictly equivalent in power to DFAs and strictly weaker than models that require stacks, tapes, or memory for counting. It provides a clean separation: if your language is regular, a feedforward neural network can model it exactly. If not—even something as simple as \(\{a^n b^n\}\)—you need memory, recurrence, or a more expressive neural architecture.

This corollary restates the classical Chomsky hierarchy \cite{chomskyhierarchy} in neural terms. While not novel in its conclusion, its utility lies in clearly delineating neural FSM expressivity limits in terms of symbolic language classes. We include it for completeness and clarity.





\section{Experiments}

\subsection{Overview and Common Setup}
\label{experimental_setup}
We empirically validate each theoretical result using neural architectures specifically designed to either mirror deterministic finite automaton (DFA) computations (via DFA-unrolled networks) or to test generalization principles through learned models.

For the experiments, we define an explicit two-state even-parity DFA over the binary alphabet $\Sigma = \{0, 1\}$, where the accepting state corresponds to strings with an even number of ones. The training data, totalling of 2000 samples, is synthetically generated by uniformly sampling binary sequences (lengths are specific to the individual experiments and are mentioned accordingly) and labeling them via the DFA’s transition logic. The dataset comprises pairs $(x, y)$ where $x \in \Sigma^T$ and $y = 1$ if the DFA accepts $x$, and $0$ otherwise. The DFA-Unrolled Network processes the sequence symbol-by-symbol, unrolling $T$ depth-wise transition layers (each with width 32), each simulating a step of the DFA’s transition function. The final output is a scalar probability (computed via a sigmoid layer) indicating acceptance by the DFA. 

Unless otherwise stated, all experiments use the Adam optimizer with a learning rate of 0.01 and full-batch training (the entire dataset per epoch) for 200 epochs. To account for variability, each experiment is repeated across five random seeds. Reported results include mean accuracy, standard deviation, and 95\% confidence intervals computed using Student's $t$-distribution. For DFA simulations, datasets were generated by uniformly sampling sequences from the DFA’s input alphabet. DFA state encodings are represented as one-hot vectors unless compact binary encodings are explicitly required, as in Lemma \hyperref[lemma_2]{2} and Corollary \hyperref[corollary_2_1]{2.1}. All experiments were run on NVIDIA GeForce RTX 4060 GPU.

\subsection{Validation of Theorem 1: Existence of FSM-Emulating Neural Networks}

To validate Theorem \hyperref[theorem_1]{1}, we construct a DFA-unrolled ReLU MLP whose architecture mirrors the DFA's transition structure via depth-wise unrolling, as described in the proof of Theorem 1 (Appendix \hyperref[proof_theorem_1]{A.1}). The DFA accepts binary strings of fixed length $T$ where the number of ones is even. Furthermore, each transition layer is itself a two-layer ReLU MLP operating on the current state and input symbol—exactly instantiating the construction described in Lemma \hyperref[lemma_1]{1}. Thus, the success of this experiment also empirically supports the claim of Lemma \hyperref[lemma_1]{1} in the context of sequential composition.

Each layer of the network simulates one transition of the DFA:
\begin{itemize}\setlength\itemsep{0em}
    \item Input symbols are one-hot encoded.
    \item Each transition layer applies the DFA's transition function deterministically.
    \item The final layer outputs 1 if the final DFA state is accepting, 0 otherwise.
\end{itemize}

Results: Table \ref{table_theorem_1} lists the accuracies achieved by DFA-unrolled MLP across five random seeds for varying input sequence lengths $T$. This deterministic simulation confirms Theorem \hyperref[theorem_1]{1}: for any DFA over finite-length strings, there exists a finite-depth ReLU neural network that computes the language indicator function exactly. Our experiment realizes this construction explicitly, providing both theoretical and empirical validation. Furthermore, it reinforces the feasibility of shallow transition networks as asserted in Lemma \hyperref[lemma_1]{1}.

\begin{table}[h]
\caption{Validation results for Theorem \hyperref[theorem_1]{1}: DFA-unrolled ReLU network simulating even-parity DFA.}
\centering
\begin{tabular}{cccc}
\toprule
\textbf{Sequence Length ($T$)} & \textbf{Mean Accuracy} & \textbf{Std Dev} & \textbf{95\% CI} \\
\midrule
{1, 2, 3, 4, 5, 6, 7} & 1.0000 & 0.0000 & $\pm$0.0000 \\
8  & 0.9975 & 0.0025 & $\pm$0.0318 \\
9  & 0.9942 & 0.0025 & $\pm$0.0318 \\
10 & 0.9817 & 0.0163 & $\pm$0.0226 \\
\bottomrule
\end{tabular}
\label{table_theorem_1}
\end{table}

\subsection{Validation of Lemma 1: Linear Separability of State Transitions}
\label{lemma_1}
To validate Lemma \hyperref[lemma_1]{1}, we train a two-layer ReLU MLP to learn the function $g: \mathbb{R}^{n + k} \rightarrow \mathbb{R}^{n}$ that realizes the transition logic $g([e_i; u_j]) = e_{\delta(q_i, s_j)}$ across varying numbers of states and symbols. Each state $q_i$ is encoded as a one-hot vector $e_i \in \mathbb{R}^n$, each symbol $s_j$ as a one-hot vector $u_j \in \mathbb{R}^k$, and the target output is $e_{\delta(q_i, s_j)}$. The input to the network is the concatenation $[e_i; u_j]$. We generate a dataset of all $n \times k$ possible inputs $[e_i; u_j] \in \mathbb{R}^{n + k}$ and corresponding outputs $e_k \in \mathbb{R}^n$.

Results: Across all tested configurations, covering 1 to 8 states ($n$) and 1 to 3 symbols ($k$) (and across five random seeds), the model consistently achieved perfect accuracy of 1.0000 (with Standard Deviation of 0.0000 and 95\% Confidence Interval of $\pm$0.0000.

This validates that DFA transitions are linearly separable and can be exactly realized using a two-layer ReLU network. The result supports Lemma \hyperref[lemma_1]{1}'s claim that the transition function over one-hot inputs is computable via a shallow MLP, further demonstrating how finite symbolic structures (like DFA transition tables) are representable within neural architectures. While the DFA-unrolled network in Theorem \hyperref[theorem_1]{1} implicitly applies a two-layer ReLU MLP per transition, this experiment isolates transition learning from sequence-level composition. It confirms Lemma \hyperref[lemma_1]{1} independently, ensuring that the transition functions are realizable without confounding effects from depth or sequential training.

\subsection{Validation of Lemma 2: State Encodings via Binary Neurons}

To validate Lemma \hyperref[lemma_2]{2}, we test whether DFA transitions can be simulated using compact binary state encodings and binary threshold units. We consider deterministic finite automata with varying numbers of states ($n = 2, 4, 8, 16, 32$), requiring $d = \lceil \log_2 n \rceil$ bits for binary encoding, and $k = 2$ input symbols. Each input to the network is a concatenation of the binary-encoded current state and a one-hot input symbol vector:
\[
[b_i; u_j] \in \{0,1\}^{d+k}.
\]
The target output is the binary representation of the next state $b_k = b_{\delta(q_i, s_j)}$.

We train a two-layer feedforward neural network with threshold-like activations: $\mathrm{sigmoid}$ activations are used during training to enable differentiability, while outputs are discretized using $\mathrm{round()}$ at inference time to simulate binary step behavior. The model is trained on all $n \cdot k$ transition pairs using 5 random seeds.

Results: Table \ref{table_lemma_2} lists the accuracies achieved by the binary transition simulator across five random seeds. This confirms that DFA transition logic can be implemented via threshold circuits over compact binary encodings, validating Lemma \hyperref[lemma_2]{2}. As the number of states increases, the performance slightly decreases due to increased learning complexity, but remains high. These results demonstrate that exponential compression of state representation is possible without sacrificing exactness for moderate-sized automata, and that neural networks with Boolean-style logic units can serve as symbolic simulators.

\begin{table}[h]
\caption{Validation results for Lemma \hyperref[lemma_2]{2}: Binary neural simulation of DFA transitions for varying state counts.}
\centering
\begin{tabular}{cccc}
\toprule
\textbf{Number of States ($n$)} & \textbf{Mean Accuracy} & \textbf{Std Dev} & \textbf{95\% CI} \\
\midrule
2   & 1.0000 & 0.0000 & $\pm$0.0000 \\
4   & 1.0000 & 0.0000 & $\pm$0.0000 \\
8   & 0.9781 & 0.0508 & $\pm$0.0228 \\
16  & 0.9202 & 0.0558 & $\pm$0.0251 \\
32  & 0.9320 & 0.0394 & $\pm$0.0386 \\
\bottomrule
\end{tabular}
\label{table_lemma_2}
\end{table}

\subsection{Validation of Theorem 2: FSM Equivalence Class Representation}

To validate Theorem \hyperref[theorem_2]{2}, we construct a DFA-unrolled ReLU network that processes entire input strings and learns an embedding that reflects equivalence class membership as defined by the Myhill-Nerode relation. Each layer of the network implements one transition of the DFA, updating a learned state representation. The final state is mapped to a continuous embedding vector, and a simple classifier is trained on top of these embeddings to predict the DFA state corresponding to each string.

Results: Table \ref{table_theorem_2} lists the accuracies achieved by the DFA-unrolled embedding network across five random seeds for varying input string lengths $T$. These results confirm Theorem \hyperref[theorem_2]{2}’s claim: by unrolling the DFA’s transition logic, the neural network learns embeddings that separate strings according to their equivalence classes. High accuracy across varying sequence lengths demonstrates that the embedding function $f_\theta$ effectively respects the equivalence relation $\equiv_L$. This provides both theoretical and empirical validation of FSM-equivalence class representation via deep networks.

\begin{table}[h]
\caption{Validation results for Theorem \hyperref[theorem_2]{2}: DFA-unrolled equivalence embeddings across varying sequence lengths.}
\centering
\begin{tabular}{cccc}
\toprule
\textbf{Sequence Length ($T$)} & \textbf{Mean Accuracy} & \textbf{Std Dev} & \textbf{95\% CI} \\
\midrule
{1, 2, 3, 4, 5, 6} & 1.0000 & 0.0000 & $\pm$0.0000 \\
7  & 0.9877 & 0.0106 & $\pm$0.0147 \\
8  & 0.9890 & 0.0123 & $\pm$0.0170 \\
9  & 0.9958 & 0.0025 & $\pm$0.0318 \\
10 & 0.9283 & 0.0887 & $\pm$0.1232 \\
\bottomrule
\end{tabular}
\label{table_theorem_2}
\end{table}

\subsection{Validation of Corollary 2.1: State Compression via Equivalence Class Embeddings}

To validate Corollary \hyperref[corollary_2_1]{2.1}, we design a neural network architecture that first simulates the DFA transition function via explicit unrolling (as in Theorem \hyperref[theorem_1]{1} and Theorem \hyperref[theorem_2]{2}), and then compresses the resulting state representations into a lower-dimensional embedding space of size $d = \lceil \log_2 n \rceil$, where $n$ is the number of DFA states.

We use mod-$n$ counter DFAs with $n \in \{2, 4, 8\}$ states, binary input alphabet $\Sigma = \{0,1\}$, and input sequence length $T = 10$. Each network consists of a linear compression layer that maps the final DFA state vector to a $\log_2(n)$-dimensional embedding and the classifier trained to predict the final DFA state from the compressed embedding. The network was trained across five random seeds per setting. We report both classification accuracy and the mean pairwise distance between compressed class centroids.

\begin{table}[h]
\caption{Validation results for Corollary \hyperref[corollary_2_1]{2.1}: DFA-unrolled compressed embeddings with increasing number of states.}
\centering
\begin{tabular}{ccccc}
\toprule
\textbf{\# States} & $d$ & \textbf{Mean Accuracy} & \textbf{Mean Class Distance} \\
\midrule
2  & 1 & $0.8639 \pm 0.0371$ (CI95: $\pm$0.1127) & $3.2595 \pm 0.9603$ (CI95: $\pm$2.9218) \\
4  & 2 & $0.9956 \pm 0.0016$ (CI95: $\pm$0.0048) & $15.8036 \pm 0.8637$ (CI95: $\pm$2.6277) \\
8  & 3 & $0.9961 \pm 0.0021$ (CI95: $\pm$0.0063) & $67.3720 \pm 16.6305$ (CI95: $\pm$50.5973) \\
\bottomrule
\end{tabular}
\label{table_corollary_2_1}
\end{table}

Results: As presented in Table \ref{table_corollary_2_1}, the DFA-unrolled networks achieved near-perfect accuracy and increasing class separation as the number of DFA states increased, even as the compressed embedding dimension scaled logarithmically with $n$. This provides empirical confirmation that DFA state spaces can be effectively compressed to $O(\log n)$ dimensions without significant loss of separability, validating Corollary \hyperref[corollary_2_1]{2.1}.

\subsection{Validation of Theorem 3: Hardness of Non-FSM Recognition}
Theorem \hyperref[theorem_3]{3} asserts that fixed-depth, finite-width feedforward neural networks cannot recognize non-regular languages such as $\mathcal{L} = \{a^n b^n \mid n \geq 1\}$, which are context-free and require unbounded memory (e.g., stack-based counting). To validate this limitation, we construct an experimental task where the network must classify whether a string belongs to $\mathcal{L}$.

Setup: Each string is over the alphabet $\Sigma = \{a, b\}$ and is one-hot encoded. Positive examples follow the pattern $a^n b^n$, while negative examples are of the form $a^n b^m$ for $m \ne n$. All sequences are padded to a maximum length of 20. The model is a fixed-size feedforward ReLU MLP, trained on examples with $n \in [1, 5]$ and tested on unseen lengths with $n \in [6, 10]$.

Results: The following results were obtained over five random seeds: \\
Mean Accuracy = 0.5080, Standard Deviation = 0.0484, and 95\% Confidence Interval = $\pm$0.0425.

The near-random performance on longer sequences confirms that the fixed-size feedforward network fails to generalize the structure of the language $a^n b^n$. This provides strong empirical support for Theorem \hyperref[theorem_3]{3}, reaffirming that non-regular language recognition requires computational capabilities beyond finite-state simulation and cannot be achieved by standard FNNs without memory mechanisms or architectural extensions.

\subsection{Validation of Corollary 3.1: Boundary of FSM-Expressive Neural Networks}

No new experiment is required for Corollary \hyperref[corollary_3_1]{3.1}, as it is empirically validated by results from Theorem \hyperref[theorem_1]{1} and Theorem \hyperref[theorem_3]{3}:
\begin{itemize}[itemsep=0pt, topsep=0pt]
    \item \textbf{Regular languages}: Theorem \hyperref[theorem_1]{1} demonstrated exact simulation of DFA behavior using DFA-unrolled ReLU MLPs, achieving perfect accuracy across input lengths.
    \item \textbf{Non-regular languages}: Theorem \hyperref[theorem_3]{3} showed that fixed-size feedforward networks failed to recognize the context-free language $L = \{ a^n b^n \}$, with accuracy degrading to random chance on longer sequences.
\end{itemize}

These results jointly confirm that finite-depth ReLU networks can express regular languages but not non-regular languages, as asserted by Corollary \hyperref[corollary_3_1]{3.1}.

\section{Conclusion}

We presented the first complete theoretical and empirical framework establishing feedforward neural networks as universal finite-state machines (N-FSMs), with a constructive and exact simulation of deterministic finite automata (DFAs). Our constructive theory shows that ReLU and threshold networks can exactly simulate deterministic finite automata (DFAs) by unrolling automaton transitions into depth-wise neural layers with formally characterized depth, width, and compression. We proved that DFA transitions are linearly separable, binary threshold activations suffice for exponential state compression, and Myhill-Nerode equivalence classes can be embedded into continuous latent spaces while preserving separability. We also formalized the expressivity boundary: fixed-depth feedforward networks cannot recognize non-regular languages, aligning neural expressivity with the Chomsky hierarchy.

Unlike prior empirical or heuristic approaches, we construct explicit DFA-unrolled neural networks that align exactly with our theoretical results. Experiments confirm DFA simulation, transition separability, binary state encoding, equivalence-class embedding, and the expressivity boundary for non-regular languages—achieving near-perfect accuracy where expected, and failure where provably necessary. These findings unify deep learning, automata theory, and neural-symbolic computation into a formal blueprint for implementing symbolic processes in continuous neural systems.

\section{Limitations}

Our results are restricted to regular languages and deterministic finite automata (DFAs). While we provide exact simulations and compression bounds for DFAs, our architectures cannot recognize non-regular languages, such as context-free languages like $\{a^n b^n\}$, as formalized and empirically validated in Theorem \hyperref[theorem_3]{3}.



\bibliographystyle{plain}
\bibliography{main}

\begin{thebibliography}{10}

\bibitem{besold2017neuralsymboliclearningreasoningsurvey}
Tarek~R. Besold, Artur d'Avila Garcez, Sebastian Bader, Howard Bowman, Pedro
  Domingos, Pascal Hitzler, Kai-Uwe Kuehnberger, Luis~C. Lamb, Daniel Lowd,
  Priscila Machado~Vieira Lima, Leo de~Penning, Gadi Pinkas, Hoifung Poon, and
  Gerson Zaverucha.
\newblock Neural-symbolic learning and reasoning: A survey and interpretation,
  2017.

\bibitem{cybenko1989approximation}
George Cybenko.
\newblock Approximation by superpositions of a sigmoidal function.
\newblock {\em Mathematics of control, signals and systems}, 2(4):303--314,
  1989.

\bibitem{garcez2019neuralsymboliccomputingeffectivemethodology}
Artur d'Avila Garcez, Marco Gori, Luis~C. Lamb, Luciano Serafini, Michael
  Spranger, and Son~N. Tran.
\newblock Neural-symbolic computing: An effective methodology for principled
  integration of machine learning and reasoning, 2019.

\bibitem{dhayalkar2025combinatorialtheorydropoutsubnetworks}
Sahil~Rajesh Dhayalkar.
\newblock A combinatorial theory of dropout: Subnetworks, graph geometry, and
  generalization, 2025.

\bibitem{dhayalkar2025geometryrelunetworksrelu}
Sahil~Rajesh Dhayalkar.
\newblock The geometry of relu networks through the relu transition graph,
  2025.

\bibitem{graves2016adaptive}
Alex Graves.
\newblock Adaptive computation time for recurrent neural networks, 2017.

\bibitem{hahn2020theoretical}
Michael Hahn.
\newblock Theoretical limitations of self-attention in neural sequence models.
\newblock {\em Transactions of the Association for Computational Linguistics},
  8:156–171, December 2020.

\bibitem{hanin2018approximatingcontinuousfunctionsrelu}
Boris Hanin and Mark Sellke.
\newblock Approximating continuous functions by relu nets of minimal width,
  2018.

\bibitem{hastad1986almost}
J~Hastad.
\newblock Almost optimal lower bounds for small depth circuits.
\newblock In {\em Proceedings of the Eighteenth Annual ACM Symposium on Theory
  of Computing}, STOC '86, page 6–20, New York, NY, USA, 1986. Association
  for Computing Machinery.

\bibitem{chomskyhierarchy}
David~G. Hays.
\newblock {\em Chomsky hierarchy}, page 210–211.
\newblock John Wiley and Sons Ltd., GBR, 2003.

\bibitem{introautomata}
John~E. Hopcroft, Rajeev Motwani, and Jeffrey~D. Ullman.
\newblock {\em Introduction to Automata Theory, Languages, and Computation (3rd
  Edition)}.
\newblock Addison-Wesley Longman Publishing Co., Inc., USA, 2006.

\bibitem{hornikmultilayer}
Kurt Hornik, Maxwell Stinchcombe, and Halbert White.
\newblock Multilayer feedforward networks are universal approximators.
\newblock {\em Neural Networks}, 2(5):359--366, 1989.

\bibitem{hupkes2020compositionality}
Dieuwke Hupkes, Verna Dankers, Mathijs Mul, and Elia Bruni.
\newblock Compositionality decomposed: how do neural networks generalise?,
  2020.

\bibitem{Johnson1984ExtensionsOL}
William~B. Johnson and Joram Lindenstrauss.
\newblock Extensions of lipschitz mappings into hilbert space.
\newblock {\em Contemporary mathematics}, 26:189--206, 1984.

\bibitem{neyshabur2015searchrealinductivebias}
Behnam Neyshabur, Ryota Tomioka, and Nathan Srebro.
\newblock In search of the real inductive bias: On the role of implicit
  regularization in deep learning, 2015.

\bibitem{rabinovich2017abstract}
Maxim Rabinovich, Mitchell Stern, and Dan Klein.
\newblock Abstract syntax networks for code generation and semantic parsing,
  2017.

\bibitem{siegelmann1995computational}
Hava~T. Siegelmann and Eduardo Sontag.
\newblock On the computational power of neural nets.
\newblock In {\em Annual Conference Computational Learning Theory}, 1992.

\bibitem{Turing1936}
Alan Turing.
\newblock On computable numbers, with an application to the
  entscheidungsproblem.
\newblock {\em Proceedings of the London Mathematical Society}, 42(1):230--265,
  1936.

\bibitem{weiss2018practical}
Gail Weiss, Yoav Goldberg, and Eran Yahav.
\newblock On the practical computational power of finite precision {RNN}s for
  language recognition.
\newblock In Iryna Gurevych and Yusuke Miyao, editors, {\em Proceedings of the
  56th Annual Meeting of the Association for Computational Linguistics (Volume
  2: Short Papers)}, pages 740--745, Melbourne, Australia, July 2018.
  Association for Computational Linguistics.

\bibitem{zhang2015learninghalfspacesneuralnetworks}
Yuchen Zhang, Jason~D. Lee, Martin~J. Wainwright, and Michael~I. Jordan.
\newblock Learning halfspaces and neural networks with random initialization,
  2015.

\end{thebibliography}

\appendix
\section{Appendix}

\subsection*{A.1 Proof of Theorem 1: Existence of FSM-Emulating Neural Networks}
\label{proof_theorem_1}
We construct such a network explicitly.

Let $n = |Q|$ and $k = |\Sigma|$. Each state $q_i \in Q$ is represented as a one-hot vector $e_i \in \mathbb{R}^n$. Each symbol $s_j \in \Sigma$ is similarly represented as a one-hot vector $u_j \in \mathbb{R}^k$.

Transition Function as a Neural Layer: The transition function $\delta: Q \times \Sigma \rightarrow Q$ can be written as a function $g: \mathbb{R}^n \times \mathbb{R}^k \rightarrow \mathbb{R}^n$ that maps $(e_i, u_j)$ to $e_{\delta(q_i, s_j)}$. Since the domain and codomain of $g$ are finite, $g$ is a discrete function on a finite domain and can therefore be implemented using a feedforward layer with ReLU activations. Specifically, for each pair $(q_i, s_j)$, we hard-code a weight matrix $W_{ij}$ such that:
\[
g(e_i, u_j) = \text{ReLU}(W_{ij} \cdot [e_i; u_j] + b_{ij}) = e_{\delta(q_i, s_j)}.
\]
We can combine all such transitions into a single layer by constructing a block-sparse weight matrix with appropriate routing logic (i.e., selector units based on the input).

Unrolling for $T$ Time Steps: We stack $T$ such transition modules, one per symbol $s_t$ in the sequence $x = (s_1, \dots, s_T)$. The output of the previous step is the current state, which along with $s_t$, determines the next state.

Let the initial state $q_0$ be encoded as $h_0 = e_0 \in \mathbb{R}^n$. Then the recursive computation is:
\[
h_1 = g(h_0, u_1), \quad
h_2 = g(h_1, u_2), \quad \dots, \quad
h_T = g(h_{T-1}, u_T).
\]

Define the output layer as a dot product $f_\theta(x) = \sigma(h_T^\top v),$ where $v \in \mathbb{R}^n$ is the indicator vector of the accepting states:
\[
v_i = 
\begin{cases}
1 & \text{if } q_i \in F, \\
0 & \text{otherwise},
\end{cases}
\]
and $\sigma$ is the identity or a step function (e.g., $\sigma(z) = \mathbb{I}[z > 0.5]$). Thus, $f_\theta(x) = 1$ iff $h_T$ corresponds to a state $q_i \in F$, i.e., the DFA accepts $x$.

Hence, the function $\phi(x)$ can be exactly computed by a feedforward network $f_\theta$ with ReLU activations and finite depth proportional to the string length $T$.

Architecture Summary: The full network is a depth-$T+1$ FNN:
\begin{itemize}
  \item Input: Sequence $x = (s_1, \dots, s_T)$, encoded as concatenated one-hot vectors in $\mathbb{R}^{Tk}$.
  \item Layers $1$ to $T$: Each simulates one $\delta$ step.
  \item Output layer: Binary classification over $F$ using final hidden state.
\end{itemize}

\subsection*{A.2 Proof of Lemma 1: Linear Separability of State Transitions}
\label{proof_lemma_1}
Let $n = |Q|$ and $k = |\Sigma|$. Fix one-hot encodings:
\[
q_i \mapsto e_i \in \mathbb{R}^n, \qquad s_j \mapsto u_j \in \mathbb{R}^k.
\]
The concatenated input $z_{ij} = [e_i; u_j] \in \mathbb{R}^{n+k}$ uniquely encodes each transition pair $(q_i, s_j)$. 

The DFA transition function $\delta$ defines a mapping $\delta: Q \times \Sigma \rightarrow Q,$ which, under encoding, corresponds to:
\[
T: \{z_{ij}\} \subset \mathbb{R}^{n+k} \rightarrow \{e_{k} \in \mathbb{R}^n \mid q_k = \delta(q_i, s_j)\}.
\]

Realizing Transition Logic via MLP: Construct an affine transformation:
\[
W \in \mathbb{R}^{n \times (n+k)}, \quad b \in \mathbb{R}^n,
\]
such that:
\[
g(z_{ij}) = \text{ReLU}(W z_{ij} + b) = e_{\delta(q_i, s_j)}.
\]

Since there are $nk$ such input-output pairs and the output space is one-hot encoded in $\mathbb{R}^n$, we can construct $nk$ affine subspaces—each projecting to a specific $e_k$. This is possible because:
\begin{itemize}
  \item The inputs $z_{ij}$ are disjoint unit vectors in the standard basis.
  \item Each corresponds to a unique target vector.
\end{itemize}

We now use a one-hidden-layer ReLU network as a lookup:
\begin{itemize}\setlength\itemsep{0em}
    \item First layer: identity matrix and appropriate shifts to isolate each \(z_{ij}\).
    \item Hidden layer: width = $nk$, with ReLU used to activate exactly one unit per input.
    \item Output layer: sum the contributions to emit the corresponding $e_k$.
\end{itemize}

For any one-hot input $z_{ij}$, the first layer computes a unique linear combination that activates only one ReLU unit in the hidden layer, say unit $\ell_{ij}$, associated to the pair $(q_i, s_j)$.

The output layer maps $\ell_{ij}$ to $e_k$ where $q_k = \delta(q_i, s_j)$. Thus:
\[
g([e_i; u_j]) = e_{\delta(q_i, s_j)}.
\]

Hence, the DFA’s transition function is linearly separable over its one-hot encoding domain and is exactly computable by a shallow ReLU network.

Network Structure:
\begin{itemize}
  \item Input: $\mathbb{R}^{n+k}$ (concatenated one-hot).
  \item Hidden: $O(nk)$ ReLU units, each uniquely mapped to a $(q_i, s_j)$ pair.
  \item Output: $\mathbb{R}^n$ (state vector).
\end{itemize}

\subsection*{A.3 Proof of Lemma 2: State Encodings via Binary Neurons}
\label{proof_lemma_2}
Let $n = |Q|$ and $k = |\Sigma|$. Assign to each state $q_i \in Q$ a unique binary code $b_i \in \{0,1\}^{d}$ where $d = \lceil \log_2 n \rceil$. Let $\mathcal{B}: Q \rightarrow \{0,1\}^d$ be this encoding.

The DFA transition function $\delta: Q \times \Sigma \rightarrow Q$ induces a function:
\[
\delta': \{0,1\}^d \times \{0,1\}^k \rightarrow \{0,1\}^d,
\]
mapping binary input $[b_i; u_j]$ to the binary encoding $b_{\delta(q_i, s_j)}$ of the next state. Since this function has finite domain size $2^d \cdot k = n \cdot k$, we can tabulate all possible inputs and outputs.

Boolean Logic Simulation: Any Boolean function \( f: \{0,1\}^m \rightarrow \{0,1\} \) can be implemented using a small threshold circuit, specifically a network of units computing:
\[
\sigma(z) = \mathbb{I}[z \geq \theta],
\]
where \( z = w^\top x \), and \(x \in \{0,1\}^m\).

This is a well-known result from circuit complexity theory:  
Every finite Boolean function is representable using a network of threshold logic units, often in depth-2 or depth-3 depending on fan-in.

Thus, for each output bit \(b_{\text{out},i} \in \{0,1\}\), we construct a Boolean circuit over input bits \([b_i; u_j]\) that decides its value.

Network Construction: Build a two-layer neural network:
\begin{itemize}
  \item Input: $d + k$ binary inputs from $b_i$ and $u_j$.
  \item Hidden layer: threshold logic gates (with weights and biases) computing relevant conjunctions/disjunctions.
  \item Output: $d$ binary values representing the bits of the next state.
\end{itemize}

Output Correctness: Each output unit implements a Boolean function:
\[
b'_\ell = f_\ell(b_i, u_j), \quad \text{for } \ell = 1, \dots, d.
\]
Thus, the output vector is precisely $b_{\delta(q_i, s_j)}$.
Hence, a binary threshold neural network can simulate the full transition function over compact binary state representations. This achieves exponential compression over one-hot encoding while preserving exact DFA behavior.

\subsection*{A.4 Proof of Theorem 2: FSM Equivalence Class Representation}
\label{proof_theorem_2}
Let $x, y \in \Sigma^{\leq T}$ be two strings of length at most $T$.

Myhill-Nerode Equivalence: Recall that $x \equiv_\mathcal{L} y$ if for all $z \in \Sigma^*$,
\[
xz \in \mathcal{L} \iff yz \in \mathcal{L}.
\]
This equivalence partitions $\Sigma^*$ into finitely many classes, each corresponding to a unique residual language. Each class corresponds to a unique state in the minimal DFA $\mathcal{A}$ accepting $\mathcal{L}$.

DFA Transition Function: Define $\hat{\delta}: \Sigma^{\leq T} \rightarrow Q$ as the extension of the DFA’s transition function over input strings:
\[
\hat{\delta}(x) = \delta(\dots \delta(q_0, s_1), \dots, s_T),
\]
where $x = s_1s_2\dots s_T$.

Thus, $x \equiv_\mathcal{L} y \iff \hat{\delta}(x) = \hat{\delta}(y)$.

Construction of Neural Mapping: Let $g: \Sigma^{\leq T} \rightarrow Q$ be the function that returns the DFA state reached after reading $x$. By Theorem \hyperref[theorem_1]{1}, $g$ is computable by a feedforward neural network $g_\theta$.

We now define a final layer that maps each state $q_i \in Q$ to a unique embedding vector $v_i \in \mathbb{R}^d$. Define:
\[
f_\theta(x) := V \cdot g_\theta(x),
\]
where $V \in \mathbb{R}^{d \times n}$ is a learnable matrix such that:
\[
V e_i = v_i, \quad \text{with } v_i \ne v_j \text{ for all } i \ne j.
\]

Injectivity and Class Consistency:
Then:
\begin{align*}
\hat{\delta}(x) = \hat{\delta}(y) &\Rightarrow g_\theta(x) = g_\theta(y) = e_i \\
&\Rightarrow f_\theta(x) = f_\theta(y) = v_i. \\
\hat{\delta}(x) \ne \hat{\delta}(y) &\Rightarrow f_\theta(x) \ne f_\theta(y).
\end{align*}

Thus, $f_\theta$ respects equivalence classes under $\equiv_\mathcal{L}$ and maps each class to a distinct vector. Hence, there exists a neural network $f_\theta$ that assigns each Myhill-Nerode equivalence class of $\mathcal{L}$ a unique embedding in $\mathbb{R}^d$.

\subsection*{A.5 Proof of Corollary 2.1: FSM State Compression via Embedding}
\label{proof_corollary_2_1}
Let $n = |Q|$ be the number of states in the minimal DFA $\mathcal{A}$. By Theorem \hyperref[theorem_2]{2}, there exists a function \( f_\theta: \Sigma^{\leq T} \rightarrow \mathbb{R}^{n} \) that maps each string \(x\) to a one-hot vector corresponding to the state reached by \(\hat{\delta}(x)\).

Let $S = \{v_1, v_2, \dots, v_n\} \subset \mathbb{R}^{n}$ be these one-hot vectors corresponding to states.

Dimension Reduction via Linear Projection: We apply a linear projection \( P: \mathbb{R}^{n} \rightarrow \mathbb{R}^d \), where \( d = \lceil \log_2 n \rceil + 1 \). Let:
\[
f_\theta'(x) := P(f_\theta(x)).
\]

We wish for the new embedding \( f_\theta'(x) \) to preserve class distinction under a norm separation criterion.

By the Johnson-Lindenstrauss lemma \cite{Johnson1984ExtensionsOL}, there exists a linear projection \( P \in \mathbb{R}^{d \times n} \) such that for any pair \(v_i, v_j \in S\), we have:
\[
\|P v_i - P v_j\|_2 \geq \epsilon > 0,
\]
for some constant \(\epsilon\), provided \(d = O(\log n)\).

The full function \( f_\theta': \Sigma^{\leq T} \to \mathbb{R}^d \) is implemented as a feedforward network:
\[
f_\theta'(x) = P \cdot f_\theta(x),
\]
where \(f_\theta(x)\) is the one-hot state vector computed from the input string as in Theorem \hyperref[theorem_2]{2}.

The projection preserves the separation of state classes because:
\[
\hat{\delta}(x) = \hat{\delta}(y) \ \Rightarrow \ f_\theta(x) = f_\theta(y) \ \Rightarrow \ f_\theta'(x) = f_\theta'(y),
\]
and
\[
\hat{\delta}(x) \ne \hat{\delta}(y) \ \Rightarrow \ f_\theta(x) \ne f_\theta(y) \ \Rightarrow \ \|f_\theta'(x) - f_\theta'(y)\| > \epsilon.
\]

Thus, there exists a function \( f_\theta' \) computable by a neural network with final output dimension \(d = O(\log n)\) that faithfully represents DFA states in a compressed latent space.

\subsection*{A.6 Proof of Theorem 3: Hardness of Non-FSM Recognition}
\label{proof_theorem_3}
Let us consider the classical non-regular, context-free language:
\[
\mathcal{L}_{CF} := \{a^n b^n \mid n \geq 1\}.
\]
This language is not regular and cannot be accepted by any DFA. We will show that a finite-depth feedforward neural network cannot compute its indicator function for unbounded \(n\).

Assume a Contradiction:
Suppose there exists a feedforward ReLU network \(f_\theta\) of fixed architecture (depth \(D\), width \(W\)) that accepts precisely the strings in \(\mathcal{L}_{CF}\), i.e.,
\[
f_\theta(x) = \phi_{CF}(x) = \mathbb{I}[x \in \mathcal{L}_{CF}],
\]
for all \(x \in \Sigma^T\), and for arbitrarily large \(T\).

Each input string \(x \in \Sigma^T\) is encoded as a concatenated one-hot matrix of dimension \(T \times |\Sigma|\), flattened into \(\mathbb{R}^{Tk}\). The network maps this input to a binary decision.

Counting Requirement of $\mathcal{L}_{CF}$:
To recognize \(a^n b^n\), the machine must count how many \(a\)’s occur (in prefix), store that count, and compare it to the number of \(b\)’s (in suffix). This requires unbounded memory: no finite-state machine or fixed-architecture FNN can maintain this count arbitrarily. It cannot “count to \(n\)” if \(n\) exceeds the effective receptive field of its depth and activations.

In particular, no matter how the weights are tuned, if \(n > D\), the network cannot distinguish between \(a^n b^n\) and \(a^n b^{n+1}\), since it lacks recurrence, memory, or adaptive depth.

Pumping Lemma for Regular Languages:
The classical pumping lemma for regular languages proves that \(\mathcal{L}_{CF}\) cannot be regular. Hence, no DFA can recognize \(\mathcal{L}_{CF}\). Since any feedforward network with fixed architecture is equivalent to a finite-state transducer (when viewed symbolically), it cannot recognize \(\mathcal{L}_{CF}\) either.

Therefore, no feedforward network of fixed architecture can compute the indicator function for \(\mathcal{L}_{CF}\) over unbounded-length strings. This proves the theorem.

\subsection*{A.7 Proof of Corollary 3.1: Boundary of FSM-Expressive Neural Networks}
\label{proof_corollary_3_1}
Part 1. Positive Expressivity

From Theorem \hyperref[theorem_1]{1} and Lemma \hyperref[lemma_1]{1}, we know that any DFA \(\mathcal{A}\) over a regular language \(\mathcal{L} = L(\mathcal{A})\) can be simulated by a feedforward neural network of finite depth and width. Thus:
\[
\mathcal{L} \in \mathcal{L}_{\text{reg}} \ \Rightarrow \ \exists f_\theta \text{ s.t. } f_\theta(x) = \mathbb{I}[x \in \mathcal{L}].
\]

Hence, the class of regular languages is fully representable by neural FSM simulators.

Part 2. Negative Expressivity

From Theorem \hyperref[theorem_3]{3}, we know that non-regular languages such as \(\mathcal{L}_{CF} = \{a^n b^n\}\) cannot be simulated by any fixed-depth feedforward neural network. This holds true for any language requiring unbounded memory, iierarchical dependencies or Nested structure (as in Dyck languages \cite{introautomata}).

Therefore:
\[
\mathcal{L} \in \mathcal{L}_{\text{non-reg}} \ \Rightarrow \ \nexists f_\theta \text{ (of finite architecture) s.t. } f_\theta(x) = \mathbb{I}[x \in \mathcal{L}] \ \forall x.
\]

Since the classes \(\mathcal{L}_{\text{reg}}\) and \(\mathcal{L}_{\text{non-reg}}\) are disjoint in the Chomsky hierarchy \cite{chomskyhierarchy}, this implies a strict boundary in neural expressivity:
\[
\text{Neural FSMs simulate exactly and only regular languages.}
\]

Conclusion:
The feedforward neural model class is expressively equivalent to deterministic finite automata and is strictly less powerful than pushdown automata or Turing machines \cite{Turing1936}.

\end{document}